\documentclass{article}

\usepackage{microtype}
\usepackage{graphicx}
\usepackage{subcaption}
\usepackage{booktabs} 

\usepackage{hyperref}

\usepackage[preprint]{icml2026}

\usepackage{amsmath}
\usepackage{amssymb}
\usepackage{mathtools}
\usepackage{amsthm}
\usepackage{paralist}
\usepackage{algorithm}
\usepackage{algorithmic}
\usepackage{fontawesome5}

\usepackage[capitalize,noabbrev]{cleveref}

\usepackage{listings}
\usepackage{xcolor}

\lstdefinestyle{promptstyle}{
    backgroundcolor=\color{gray!10},
    basicstyle=\ttfamily\footnotesize,
    breaklines=true,
    breakatwhitespace=false,
    frame=single,
    framesep=5pt,
    xleftmargin=10pt,
    xrightmargin=10pt,
    numbers=none,
    showstringspaces=false,
    columns=fullflexible,
    keepspaces=true,
}

\theoremstyle{plain}

\theoremstyle{definition}

\theoremstyle{remark}

\usepackage[textsize=tiny]{todonotes}

\icmltitlerunning{Error Taxonomy-Guided Prompt Optimization}

\begin{document}

\twocolumn[
  \icmltitle{Error Taxonomy-Guided Prompt Optimization}



  \icmlsetsymbol{equal}{*}

  \begin{icmlauthorlist}
    \icmlauthor{Mayank Singh}{uoa}
    \icmlauthor{Vikas Yadav}{svcenw}
    \icmlauthor{Eduardo Blanco}{uoa}
  \end{icmlauthorlist}

  \icmlaffiliation{uoa}{University of Arizona}
  \icmlaffiliation{svcenw}{ServiceNow}

  \icmlcorrespondingauthor{Mayank Singh}{mayanks43@arizona.edu}

  \icmlkeywords{
    Large Language Models,
    Automatic Prompt Optimization,
    Natural Language Feedback,
    Error Analysis
  }

  \vskip 0.3in
]



\printAffiliationsAndNotice{}  

\begin{abstract}
Automatic Prompt Optimization (APO) is a powerful approach
for extracting performance from large language models
without modifying their weights.
Many existing methods rely on trial-and-error, testing different prompts
or in-context examples until a good configuration emerges,
often consuming substantial compute.
Recently, natural language feedback derived from execution logs
has shown promise as a way to identify how prompts can be improved.
However, most prior approaches operate in a bottom-up manner,
iteratively adjusting the prompt based on feedback from individual problems,
which can cause them to lose the global perspective.
In this work, we propose Error Taxonomy-Guided Prompt Optimization (ETGPO),
a prompt optimization algorithm that adopts a top-down approach.
ETGPO focuses on the global failure landscape by collecting model errors,
categorizing them into a taxonomy, and augmenting the prompt with guidance
targeting the most frequent failure modes.
Across multiple benchmarks spanning mathematics, question answering,
and logical reasoning, ETGPO achieves accuracy that is comparable to
or better than state-of-the-art methods, while requiring roughly one third
of the optimization-phase token usage and evaluation budget.
\end{abstract}

\section{Introduction}
\label{s:introduction}


Large language models (LLMs) are sensitive to prompt wording, structure,
and examples, making prompt design an effective mechanism for improving
task performance without modifying model weights.
However, manual prompt design requires expertise, trial-and-error,
and effort.
This has motivated a growing body of work on Automatic Prompt Optimization (APO),
which treats prompt design as an optimization problem over discrete natural-language strings.
Early approaches include APE~\cite{ape},
which framed prompt design as a black-box optimization problem;
OPRO~\cite{opro}, which showed that LLMs can optimize
prompts by learning from past attempts; and ProTeGi~\cite{protegi}, which introduced
feedback-driven refinement using natural language critiques.

\begin{figure}[t]
\centering
\begin{tikzpicture}[
    box/.style={
        rectangle,
        draw=black!70,
        rounded corners,
        minimum width=5.5cm,
        minimum height=1.1cm,
        align=center,
        font=\small
    },
    arrow/.style={
        ->,
        thick,
        black!70
    },
    label/.style={
        font=\footnotesize,
        text=gray
    }
]

\node[box, fill=blue!15] (collect) at (0, 0) {\faSearch\ \textbf{Error Collection}\\$K$ runs on validation set};
\node[box, fill=orange!15] (taxonomy) at (0, -1.9) {\faLayerGroup\ \textbf{Error Taxonomy Creation}\\Categorize failed traces};
\node[box, fill=green!15] (select) at (0, -3.8) {\faFilter\ \textbf{Category Selection}\\Select top $G$ categories};
\node[box, fill=purple!15] (guidance) at (0, -5.7) {\faLightbulb\ \textbf{Guidance Generation}\\Generate actionable guidance};

\node[box, fill=gray!15] (output) at (0, -7.6) {\faRocket\ \textbf{Optimized Prompt}};

\draw[arrow] (collect) -- (taxonomy) node[midway, right, label] {\textcolor{red!70}{\faBug} failed traces};
\draw[arrow] (taxonomy) -- (select) node[midway, right, label] {\textcolor{orange!70}{\faSitemap} error taxonomy};
\draw[arrow] (select) -- (guidance) node[midway, right, label] {\textcolor{green!70}{\faCheckSquare} selected categories};
\draw[arrow] (guidance) -- (output) node[midway, right, label] {\textcolor{purple!70}{\faMagic} final prompt};

\end{tikzpicture}
\caption{
    Overview of ETGPO. We first collect errors from failed traces
    and then organize them into a taxonomy.
    After filtering to the most prevalent error categories,
    we generate actionable guidance for each and combine them into the final prompt.
    This entire process is automated end-to-end.
}
\label{fig:method_overview}
\end{figure}
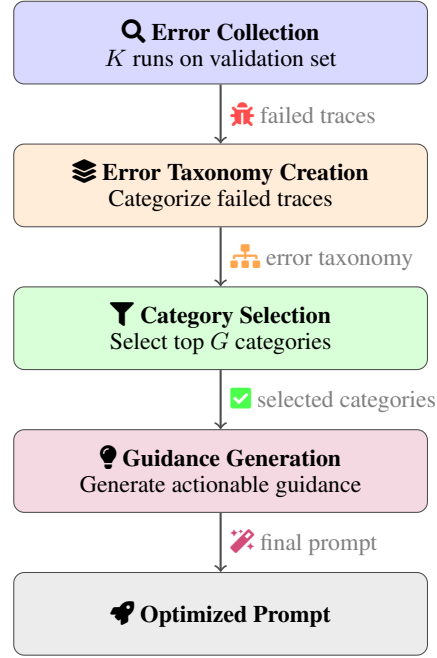

Since then, diverse approaches have emerged: evolutionary methods that evolve
prompts through mutation and combination~\cite{evoprompt, promptbreeder},
history-informed methods that use Bayesian optimization to guide
search~\cite{mipro}, and further feedback-driven methods that refine prompts
based on execution errors~\cite{textgrad, promptwizard, gepa}.

We focus on feedback-driven approaches, which have shown strong empirical gains
but operate bottom-up: they collect errors on small batches and immediately
edit the prompt, risking overfitting to specific examples. To mitigate this,
they allocate substantial budget to repeatedly evaluating each candidate on a
validation set \cite{gepa}.

To address these issues, we propose \textbf{Error Taxonomy-Guided Prompt
Optimization (ETGPO)}, a top-down APO method that explicitly models the
dataset-level error landscape through an error taxonomy. The taxonomy
organizes errors found in the execution traces of a base prompt into
error categories, along with their prevalence statistics, providing an
interpretable view of the errors the LLM makes on the domain of interest.
This structure allows us to generate targeted guidance only for the most prevalent
error categories. By filtering out long-tail errors, we ensure that prompt
edits are more likely to generalize beyond specific problems, reducing the
need for iterative validation and improving overall efficiency.

In summary, our contributions are \footnote{
    Code at: \url{https://github.com/mayanks43/etgpo}
}:
\begin{compactenum}
\item We propose ETGPO, a top-down approach to APO that builds a global
error taxonomy and generates guidance for prevalent failure modes.
ETGPO is independent of the task and choice of LLM,
while also providing an interpretable view of the target LLM's weaknesses.
\item We show that ETGPO matches or outperforms state-of-the-art methods
on benchmarks spanning mathematics, question answering, and logical reasoning.
\item We achieve this at roughly one third of the optimization cost,
demonstrating that a simpler approach can replace expensive iterative refinement.
\end{compactenum}

\section{Related Work}
\label{s:related_work}

There exist many APO methods in the literature, and they can be classified in
several different ways.
For our purposes, we group them into three categories based on how new
candidates are proposed: feedback-driven, evolutionary, and
history-informed methods, and discuss representative approaches within
each category.

Feedback-driven approaches iteratively refine a prompt by analyzing
its performance on examples and using that critique to guide edits.
ProTeGi~\cite{protegi} iteratively collects errors on mini-batches of training examples,
uses an LLM to summarize what went wrong, and edits the prompt accordingly.
It maintains multiple candidates via beam search, selecting the best
by validation performance.
TextGrad~\cite{textgrad} frames prompt optimization as gradient descent, where an LLM
generates critiques of outputs and another LLM edits the prompt to
address them. It follows a single optimization trajectory rather than
maintaining multiple candidates via beam search.
PromptWizard~\cite{promptwizard} also uses an LLM to analyze reasoning failures,
identify prompt weaknesses, and improve the prompt.
However, unlike TextGrad, it additionally collects in-context examples
from successful runs, jointly optimizing the prompt and in-context examples.

Evolutionary methods maintain multiple prompt candidates, evolving
them through combination and mutation while discarding weaker candidates.
EvoPrompt~\cite{evoprompt} maintains a pool of prompt candidates and uses an LLM to
combine or rewrite them, filtering to top-k after each round.
PromptBreeder~\cite{promptbreeder}
extends EvoPrompt by evolving both the task prompts
and the prompts used to mutate them,
so the system learns better mutation strategies over time.
GEPA~\cite{gepa} is a hybrid method that mutates and combines prompts based on
execution feedback while maintaining a diverse population where each
candidate excels on a different subset of examples, with the best
prompt selected by overall coverage.

History-informed methods improve prompts iteratively, using the
trajectory of past attempts and their scores to inform new candidates.
OPRO~\cite{opro} uses an LLM to propose new prompts while showing it the
history of past prompts and their scores, letting it learn from previous
attempts.
APE~\cite{ape} uses an LLM to generate prompt candidates and iteratively
refines them by paraphrasing top performers, treating successful prompts
as implicit historical guidance.
MIPRO~\cite{mipro} also uses an LLM to generate prompt candidates, but additionally
optimizes in-context examples from successful runs, using Bayesian
optimization to select promising combinations.

ETGPO, our method, is feedback-driven but takes a different approach than
existing methods. Rather than maintaining multiple prompt candidates like
evolutionary and history-informed methods, or iteratively editing based on
individual errors like other feedback-driven methods, ETGPO builds a global
error taxonomy and generates targeted guidance for the most prevalent failure
modes. This top-down approach enables surgical, high-impact interventions,
reducing the need for repeated candidate evaluations and lowering overall
optimization cost.

\section{Method}
\label{s:method}

Before describing our method, we introduce the problem setting and some necessary
definitions.

We consider a dataset of problems, each with a ground-truth answer.
We are also given a base system prompt, which is used to query a large language
model together with the problem text to produce a predicted answer.
The goal of automatic prompt optimization is to improve this base system prompt
by appending additional text such that the accuracy of the model's responses,
measured against the ground-truth answers, improves over the dataset under
consideration overall.

We now introduce the definitions.
First, we distinguish between two types of LLMs used in our method.
The \emph{backbone LLM} is the model used to solve problems in the dataset under
consideration (i.e., generate reasoning traces and answers).
The \emph{optimizer LLM} is the model used to improve the system prompt that is
passed to the backbone LLM.

Second, we define \emph{failed traces} as the reasoning traces of problems in the
dataset for which the backbone LLM fails to produce the correct answer.
Third, we define an \emph{error taxonomy} as a categorization of the errors
identified in the failed traces, along with their prevalence statistics.
Fourth, we define \emph{guidance} as a set of instructional text designed to
prevent the kinds of errors described by the error taxonomy.

Next, we describe ETGPO, our method. It consists of four steps:
(1) error collection,
(2) error taxonomy creation,
(3) error category selection, and
(4) guidance generation.
We describe each step in detail below:

\subsection{Error Collection}
In the first step, we take the validation set of the dataset at hand together
with the base system prompt and query the backbone LLM on this set of problems.
We then collect the failed traces, along with the problem statements,
correct answers, and predicted answers.
Rather than making a single pass over the validation set,
we perform \(K\) runs to capture the variety of errors that can arise
from the same problem due to the LLM's stochastic nature.

\begin{algorithm}[t]
  \caption{Error Taxonomy-Guided Prompt Optimization}
  \label{alg:taxonomy_optimization}
  \begin{algorithmic}
\STATE {\bfseries Input:} Validation set $V$, base prompt $P$, backbone LLM, optimizer LLM
\STATE {\bfseries Parameters:} $K$ (collection runs), $B$ (batch size), $G$ (max error categories)
\STATE {\bfseries Output:} Optimized prompt $P^*$
\STATE
\STATE \textit{// Step 1: Error Collection}
\STATE $F \leftarrow \emptyset$
\FOR{$k = 1$ {\bfseries to} $K$}
    \FOR{each problem in $V$}
        \STATE Query backbone LLM with $P$ and problem
        \IF{prediction is incorrect}
            \STATE Add (problem, trace, prediction) to $F$
        \ENDIF
    \ENDFOR
\ENDFOR
\STATE
\STATE \textit{// Step 2: Error Taxonomy Creation}
\STATE $C \leftarrow \emptyset$
\FOR{each batch of $B$ traces in $F$}
    \STATE Use optimizer LLM to identify error categories
    \STATE Update $C$ with new or existing error categories
\ENDFOR
\STATE
\STATE \textit{// Step 3: Error Category Selection}
\STATE Filter $C$ to error categories with $\geq 2$ problems
\STATE Sort $C$ by failure count (descending)
\STATE $S \leftarrow$ top $G$ error categories from $C$
\STATE
\STATE \textit{// Step 4: Guidance Generation}
\STATE Use optimizer LLM to generate guidance for $S$
\STATE $P^* \leftarrow P$ + preamble + guidance texts
\STATE \textbf{return} $P^*$
\end{algorithmic}

\end{algorithm}

\subsection{Error Taxonomy Creation}

We next create a taxonomy of the errors found in the failed traces.
For each failed trace, we prompt the optimizer LLM to analyze the problem,
given the problem statement, the correct answer, the model's reasoning trace,
and the model's predicted answer.
The optimizer LLM is asked to identify where the reasoning first goes wrong,
determine the nature of the error, and explain why this error led to an incorrect
answer.
Based on this analysis, the LLM assigns the trace to an error category.

\begin{figure*}[t]
\centering
\input{figures/optimized_prompt}
\caption{
  Comparison between the original system prompt and the ETGPO-optimized
  prompt for AIME, using GPT-4.1-mini as the backbone model and GPT-4.1
  as the optimizer model. The complete optimized prompt has 10 blocks,
  of which only the first block and a portion of the second are shown.
}
\label{fig:optimized_prompt}
\end{figure*}

In practice, the number of failed traces can be large, which risks exceeding the
context length of the optimizer LLM.
To address this, we perform taxonomy creation in a batched manner.
In the first call, we provide the optimizer LLM with a batch of \(B\) failed traces,
along with their associated information as described above, and ask it to group them
into categories based on the error analysis.
In response, the LLM assigns each trace to a newly created error category and
produces auxiliary information for each category, including a summary, a description,
an example, the error type, and an explanation of why the error leads to an
incorrect answer.
Each error category is required to be self-contained and understandable
without reference to the original problems, so that the resulting guidance
generalizes beyond the specific examples encountered during optimization.

In subsequent calls, we repeat this process with the additional constraint that
the LLM should reuse existing error categories whenever possible.
Accordingly, in these calls, we also provide the LLM with the current set of error
categories, along with metadata such as the number of traces assigned to each
category so far.
At the end of this process, we obtain a set of error categories that form the error
taxonomy, together with their prevalence statistics.
The full prompt for this step is provided in \cref{a:appendix_1}.

\begin{table*}[t]
  \centering
  \caption{
    Results (mean accuracy with confidence intervals across all runs) obtained using basic chain-of-thought
    and three automatic prompt optimization methods, with \textbf{GPT-4.1-mini} used as the backbone model
    and GPT-4.1 used as the optimizer.
    ETGPO (our method) outperforms prior methods on average and on all but one benchmark.
    Note: \textbf{Bold} = best; \underline{underline} = second-best.
  }
  \begin{tabular}{@{} lrrrrrrr @{}}
\toprule
\textbf{Method}
& \multicolumn{2}{c}{\bf Math Reasoning}
& \bf General
& \bf Multi-hop
& \multicolumn{2}{c}{\bf Logical Reasoning}
& \bf Avg. \\
\cmidrule(lr){2-3} \cmidrule(lr){6-7}
& \bf AIME
& \bf HMMT
& \bf MMLU-Pro
& \bf Musique
& \bf AR-LSAT
& \bf FOLIO
& \\
\midrule

Chain of Thought
& 47.08 ± 1.43
& 47.14 ± 1.23
& 78.82 ± 0.30
& 74.25 ± 0.63
& 74.35 ± 0.91
& 75.09 ± 0.67
& 66.12 \\

MIPROv2
& \underline{47.66 ± 1.22}
& 45.89 ± 1.31
& 78.37 ± 0.42
& 75.20 ± 0.60
& \underline{76.36 ± 1.11}
& \underline{79.71 ± 0.92}
& 67.20 \\

GEPA
& \textbf{49.06 ± 1.51}
& \underline{47.86 ± 1.38}
& \textbf{79.55 ± 0.58}
& \underline{76.35 ± 0.51}
& 75.60 ± 0.98
& 77.83 ± 0.90
& \underline{67.71} \\

ETGPO (Ours)
& \textbf{49.06 ± 1.36}
& \textbf{50.47 ± 1.28}
& \underline{79.40 ± 0.66}
& \textbf{77.30 ± 0.45}
& \textbf{76.93 ± 0.76}
& \textbf{81.34 ± 0.60}
& \textbf{69.08} \\

\bottomrule
\end{tabular}

  \label{t:accuracy_gpt41mini}
\end{table*}

\begin{table*}[t]
  \centering
  \caption{
    Optimization-phase token usage (total input plus output tokens, truncated to thousands)
    for three automatic prompt optimization methods, with \textbf{GPT-4.1-mini} used as the backbone
    model and GPT-4.1 used as the optimizer.
    The $\Delta$ column denotes the ratio of each method's average token usage to that of ETGPO.
    ETGPO (our method) consistently uses the fewest tokens across all benchmarks.\\
    Note: \textbf{Bold} = lowest; \underline{underline} = second-lowest.
  }
  \begin{tabular}{@{} lrrrrrr rr @{}}
\toprule
\textbf{Method}
& \textbf{AIME}
& \textbf{HMMT}
& \textbf{MMLU-Pro}
& \textbf{Musique}
& \textbf{AR-LSAT}
& \textbf{FOLIO}
& \textbf{Avg.}
& \textbf{$\Delta$} \\
\midrule

MIPROv2
& 14{,}332
& 11{,}629
& 3{,}859
& 23{,}735
& 6{,}954
& 2{,}275
& 10{,}464
& 4.6$\times$ \\

GEPA
& \underline{12{,}495}
& \underline{10{,}249}
& \underline{2{,}623}
& \underline{6{,}093}
& \underline{5{,}144}
& \underline{2{,}320}
& \underline{6{,}487}
& 2.8$\times$ \\

ETGPO (Ours)
& \textbf{4{,}377}
& \textbf{4{,}025}
& \textbf{595}
& \textbf{2{,}849}
& \textbf{1{,}372}
& \textbf{453}
& \textbf{2{,}279}
& --- \\

\bottomrule
\end{tabular}

  \label{t:tokens_gpt41mini}
\end{table*}

\subsection{Error Category Selection}
After building the error taxonomy, we select a subset of error categories for
generating guidance that helps the backbone LLM avoid such errors.
Using all error categories is impractical for several reasons:
it would result in an overly long prompt,
and many categories are rare or occur in only a single problem.
Focusing on the most common errors is likely to have greater impact.

For these reasons, we focus on selecting the most important error categories.
For each error category, we obtain two types of statistics from the previous step:
the \emph{problem count}, defined as the number of unique problems exhibiting the
error, and the \emph{failure count}, defined as the number of failed traces
associated with the category.
To avoid optimizing for error categories that are overly specific to individual
problems, we filter out categories associated with only a single problem.
After this filtering step, we select the top \(G\) error categories
by failure count for guidance generation in the next step.

\subsection{Guidance Generation}
We next generate actionable guidance text for each selected error category.
We prompt the optimizer LLM with information about each category, including its
description, an example, and an explanation of why it leads to an incorrect
answer, and ask it to produce guidance text for each category.
The LLM is tasked with generating actionable advice together with concrete
incorrect and correct examples.
In addition, the LLM generates a preamble that introduces the guidance.
All guidance texts for the selected error categories, along with the preamble, are
generated in a single LLM call.
After receiving the response from the LLM, the preamble and guidance texts are
appended to the original system prompt.
The output of this step is the final improved system prompt.
The full prompt used to query the optimizer LLM
in this step is provided in \cref{a:appendix_2}.
\cref{fig:optimized_prompt}
shows an example prompt generated by our method. Additional ETGPO-generated
prompts are provided in \cref{a:appendix_3}.

We experiment with two styles of guidance: a short style, consisting of one to
two sentences per error category, and a detailed style, which includes a full
description, actionable advice, and incorrect and correct examples.
We later perform an ablation study comparing these two styles.

\section{Evaluation}
\label{s:evaluation}

\begin{table*}[t]
  \centering
  \caption{
    Results (mean accuracy with confidence intervals across all runs) obtained using basic chain-of-thought
    and three automatic prompt optimization methods, with \textbf{DeepSeek-V3.1} used as the backbone model
    and GPT-4.1 used as the optimizer.
    ETGPO (our method) outperforms prior methods on average and on three of six benchmarks.
    Note: \textbf{Bold} = best; \underline{underline} = second-best.
  }
  \begin{tabular}{@{} lrrrrrrr @{}}
\toprule
\textbf{Method}
& \multicolumn{2}{c}{\bf Math Reasoning}
& \bf General
& \bf Multi-hop
& \multicolumn{2}{c}{\bf Logical Reasoning}
& \bf Avg. \\
\cmidrule(lr){2-3} \cmidrule(lr){6-7}
& \bf AIME
& \bf HMMT
& \bf MMLU-Pro
& \bf Musique
& \bf AR-LSAT
& \bf FOLIO
& \\
\midrule

Chain of Thought
& 65.52 ± 1.25
& 53.65 ± 1.28
& 82.74 ± 0.53
& \underline{76.92 ± 0.39}
& 90.65 ± 0.61
& 75.58 ± 0.88
& 74.18 \\

MIPROv2
& \underline{68.23 ± 1.25}
& \textbf{60.68 ± 1.51}
& \underline{83.73 ± 0.54}
& 76.80 ± 0.61
& \textbf{91.66 ± 0.68}
& \underline{82.33 ± 0.57}
& \underline{77.24} \\

GEPA
& 64.48 ± 1.33
& 57.55 ± 1.30
& \textbf{83.76 ± 0.66}
& 76.38 ± 0.55
& 90.92 ± 0.57
& \textbf{82.45 ± 0.40}
& 75.92 \\

ETGPO (Ours)
& \textbf{69.74 ± 1.33}
& \underline{58.65 ± 1.62}
& 83.65 ± 0.52
& \textbf{79.03 ± 0.72}
& \underline{91.44 ± 0.52}
& \textbf{82.45 ± 0.93}
& \textbf{77.49} \\

\bottomrule
\end{tabular}

  \label{t:accuracy_deepseekv31}
\end{table*}

\begin{table*}[t]
  \centering
  \caption{
    Optimization-phase token usage (total input plus output tokens, truncated to thousands)
    for three automatic prompt optimization methods, with \textbf{DeepSeek-V3.1} used as the backbone
    model and GPT-4.1 used as the optimizer.
    The $\Delta$ column denotes the ratio of each method's average token usage to that of ETGPO.
    ETGPO (our method) consistently uses the fewest tokens across all benchmarks.\\
    Note: \textbf{Bold} = lowest; \underline{underline} = second-lowest.
  }
  \begin{tabular}{@{} lrrrrrr rr @{}}
\toprule
\textbf{Method}
& \textbf{AIME}
& \textbf{HMMT}
& \textbf{MMLU-Pro}
& \textbf{Musique}
& \textbf{AR-LSAT}
& \textbf{FOLIO}
& \textbf{Avg.}
& \textbf{$\Delta$} \\
\midrule

MIPROv2
& 17{,}400
& 17{,}027
& 4{,}785
& 26{,}096
& 10{,}735
& 3{,}276
& 13{,}220
& 6.5$\times$ \\

GEPA
& \underline{9{,}899}
& \underline{9{,}923}
& \underline{3{,}300}
& \underline{6{,}449}
& \underline{5{,}523}
& \underline{2{,}550}
& \underline{6{,}274}
& 3.1$\times$ \\

ETGPO (Ours)
& \textbf{3{,}098}
& \textbf{3{,}295}
& \textbf{778}
& \textbf{3{,}063}
& \textbf{1{,}378}
& \textbf{507}
& \textbf{2{,}020}
& --- \\

\bottomrule
\end{tabular}

  \label{t:tokens_deepseekv31}
\end{table*}

In this section, we discuss the evaluation setup and results for our method on various
datasets.

\subsection{Datasets}

We evaluate our method on six datasets spanning multiple domains:
two mathematics datasets (AIME, HMMT),
general question answering (MMLU-Pro),
multi-hop reasoning (MuSiQue),
and logical and analytical reasoning (AR-LSAT, FOLIO).
We describe them below:

\textbf{AIME}~\cite{aime}:
The American Invitational Mathematics Examination (AIME) is a challenging
high-school mathematics competition that tests advanced mathematical skills.
We use questions from AIME to evaluate the mathematical reasoning abilities of
LLMs.
In our experiments, we use 90 problems from the 2022--2024 editions as the
validation set and 30 problems from the 2025 edition as the test set.

\textbf{HMMT}~\cite{hmmt}:
The Harvard--MIT Mathematics Tournament (HMMT) is another challenging
high-school mathematics competition.
For our experiments, we use 90 problems from the February 2023, 2024, and 2025
editions as the validation set, and 30 problems from the November 2025 edition as
the test set.
We use the publicly available datasets released
by MathArena~\cite{matharena}.

\textbf{MMLU-Pro}~\cite{mmlupro}:
MMLU-Pro is a more challenging variant of the MMLU benchmark~\cite{mmlu}, a general-domain
question answering dataset.
Compared to the original benchmark, which provides four answer choices per
question, MMLU-Pro includes ten options per question.
Since the original validation set contains only 70 problems, we instead create
larger validation and test sets by partitioning the full test set into two
subsets.
We then sample 100 problems from the validation subset and 500 problems from the
test subset to form the validation and test sets, respectively.

\textbf{MuSiQue}~\cite{musique}:
MuSiQue is a multi-hop reasoning dataset whose questions require answering
multiple implicit subquestions to arrive at the correct answer.
The dataset is explicitly constructed to avoid questions that can be solved by
answering only a single subquestion, thereby requiring genuine multi-hop
reasoning.
Due to the large size of the dataset, we randomly sample 160 questions from the
training split to form the validation set and 500 questions from the validation
split to form the test set.

\textbf{AR-LSAT}~\cite{ar-lsat}:
AR-LSAT is an analytical reasoning dataset derived from questions in the Law
School Admission Test (LSAT).
We use a random subset of 100 questions from the validation split as the
validation set and include all 230 questions from the test split as the test set.

\textbf{FOLIO}~\cite{folio}:
FOLIO is a dataset designed to test complex logical reasoning abilities.
Each instance consists of premises and a conclusion, and the task is to
determine whether the conclusion logically follows from the premises.
For our experiments, we use a random subset of 100 questions from the training
split as the validation set and all 203 questions from the validation
split as the test set.

\subsection{Baselines}
We include Chain of Thought~\cite{chain-of-thought}
as our primary baseline, since the other APO baselines
in this paper improve upon its prompt.
We also include two state-of-the-art APO methods for comparison:
GEPA~\cite{gepa} and MIPRO~\cite{mipro}.
We use the DSPy~\cite{dspy}
implementations of GEPA and MIPRO; in DSPy, MIPRO is implemented as MIPROv2.
For both GEPA and MIPRO, we use the heavy optimization mode, which uses the
largest number of evaluations to search for the best prompt.
Our goal is to compare against the strongest versions of these methods available.

\subsection{Experimental Results}
\paragraph{Accuracy results}
We evaluate our method and the baselines on the test sets of the listed datasets.
A subset of the results is shown in \cref{t:accuracy_gpt41mini}
(backbone LLM: GPT-4.1-mini~\cite{gpt4.1-mini}) and \cref{t:accuracy_deepseekv31}
(backbone LLM: DeepSeek-V3.1~\cite{deepseek-V3.1}).
All reported numbers correspond to the accuracy achieved by the final prompts
produced by each method when evaluated on the corresponding test sets.

For all APO methods reported in these tables, we use GPT-4.1~\cite{gpt4.1} as the optimizer LLM.
To reduce computational cost, we run each APO method only once per dataset, but
evaluate the resulting prompt multiple times.
For the smaller mathematics datasets, we perform 64 runs per problem.
For the larger datasets, we use fewer runs per problem, specifically 8 runs for
MuSiQue and MMLU-Pro, and 16 runs for AR-LSAT and FOLIO.
We use a high number of runs per prompt to obtain tighter confidence intervals
and stronger statistical reliability.
Following recent work~\cite{llm-as-a-judge}, we use LLM-based answer equivalence
checking: a prediction is correct if it exactly matches the ground truth or,
when exact match fails, is judged equivalent by a strong LLM (GPT-5~\cite{gpt5}).

\begin{figure*}[t]
  \centering
  \begin{subfigure}[b]{0.48\textwidth}
    \includegraphics[width=\textwidth]{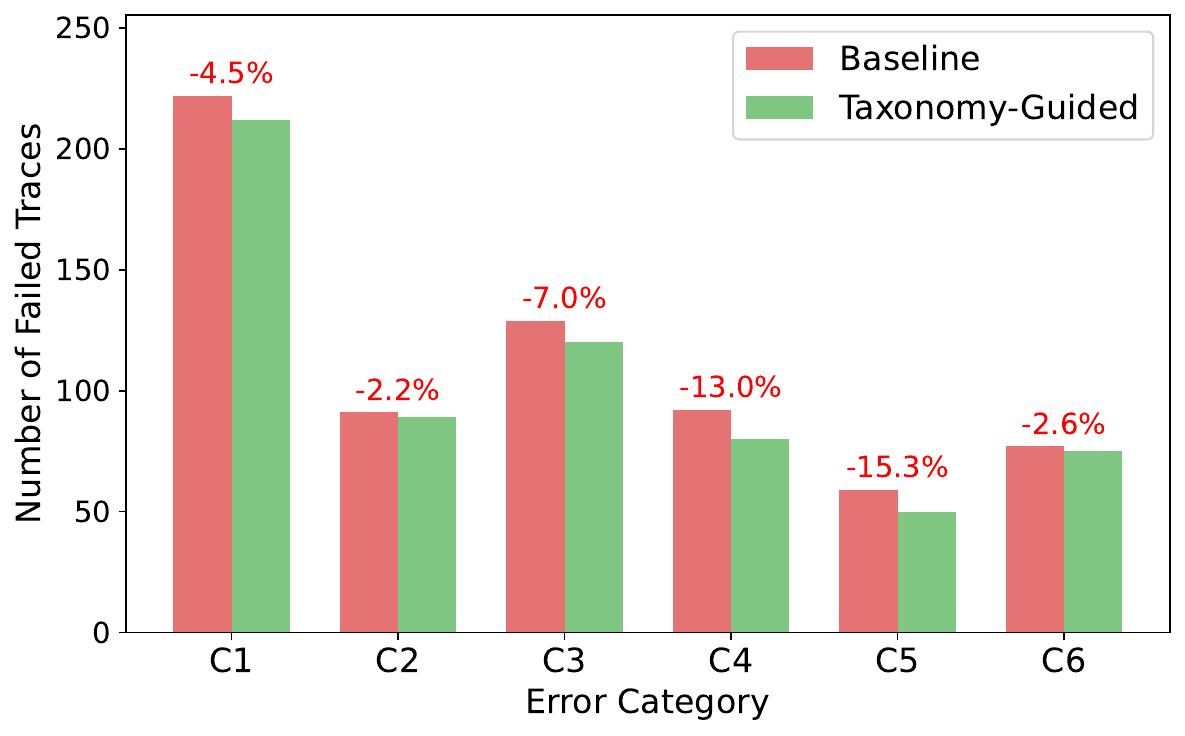}
    \caption{}
    \label{fig:aimo_failure_reduction}
  \end{subfigure}
  \hfill
  \begin{subfigure}[b]{0.48\textwidth}
    \includegraphics[width=\textwidth]{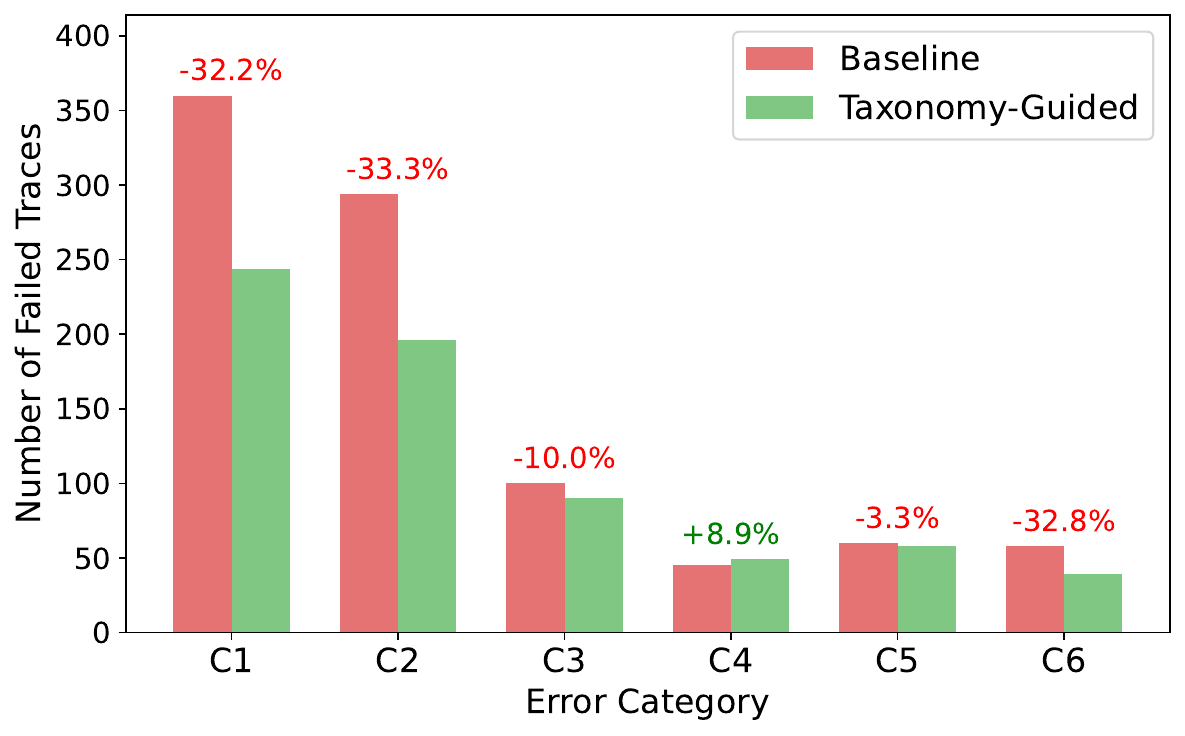}
    \caption{}
    \label{fig:folio_failure_reduction}
  \end{subfigure}
  \caption{
    Count of failed traces before and after optimization with ETGPO for the
    top 6 error categories from the error taxonomy on the validation sets
    of (a) AIME and (b) FOLIO. All but one category shows fewer failed traces,
    supporting our hypothesis that targeted guidance reduces error prevalence.
  }
  \label{fig:failure_reduction}
\end{figure*}

For our method, we use the same hyperparameter values across all datasets, chosen
based on a hyperparameter search conducted on the AIME validation set.
To reduce computational cost, we do not perform a separate hyperparameter search
for each dataset.
Specifically, we set \(K\), the number of error collection runs, to 5; \(G\), the
number of selected error categories, to 10; and use detailed guidance texts rather
than the short variant.
We do not perform a search over the batch size parameter \(B\) and instead use
the default value of 6.
Tuning these hyperparameters on a per-dataset basis could further improve performance.

We now discuss the results shown in \cref{t:accuracy_gpt41mini} and
\cref{t:accuracy_deepseekv31}.
Our method achieves the best average performance across both tables, consistently
matching or outperforming other APO methods.
On individual datasets, it performs best on 5 out of 6 with GPT-4.1-mini as the
backbone LLM, and 3 out of 6 with DeepSeek-V3.1.

In terms of absolute improvements on individual datasets, the largest gains are
observed on FOLIO for both backbone models, with improvements of approximately
6--7 percentage points in accuracy.
AIME, HMMT, and MuSiQue also show strong gains, ranging from roughly 2 to 5
percentage points across the two backbone models consistently.
In contrast, gains on MMLU-Pro are modest, which we attribute to the fact that
these questions often require more specific domain knowledge that is less likely
to be improved through general guidance alone; retrieval-based approaches may be more
suitable in this setting.
AR-LSAT shows a strong improvement of approximately 2.5 percentage points when
using GPT-4.1-mini as the backbone LLM.
However, performance on AR-LSAT appears to be saturated for DeepSeek-V3.1, with
accuracies approaching 90\%, leaving limited room for further improvement.
In this regime, the execution logs provide little additional signal from which to
extract useful guidance, resulting in marginal gains.

\begin{table}[t]
  \centering
  \caption{
    Effect of varying \(K\), the number of error collection runs, on final
    prompt accuracy for the AIME and FOLIO validation sets. Accuracy saturates
    at different values of \(K\), suggesting the optimal number of runs is
    dataset-dependent.
  }
  \begin{tabular}{@{} lcc @{}}
\toprule
\textbf{K} & \bf AIME & \bf FOLIO \\
\midrule

Baseline
& 52.35 ± 0.91
& 72.50 ± 0.08 \\

1
& 52.61 ± 0.59
& \textbf{79.70 ± 0.37} \\

5
& \textbf{53.72 ± 0.28}
& \underline{79.53 ± 0.53} \\

10
& \underline{53.43 ± 1.06}
& 78.32 ± 0.38 \\

\bottomrule
\end{tabular}

  \label{t:hyperparameter_k}
\end{table}

\paragraph{Token usage results}
In \cref{t:tokens_gpt41mini} and \cref{t:tokens_deepseekv31}, we report the total
number of input and output tokens consumed during the optimization phase of each
APO method evaluated, including our method, with all values reported in thousands
and truncated for clarity.
In all cases, the reported token counts correspond to the sum of tokens consumed
by the backbone model and, when applicable, the optimizer LLM.

Across both backbone LLMs and all datasets, our method consistently uses the
fewest tokens during the optimization phase.
In particular, the total token usage of our method is approximately one third of
that of the next most expensive method, GEPA.
These results demonstrate that our method achieves performance that is on par
with or better than state-of-the-art APO methods, while being substantially more
token-efficient during optimization.

\begin{table*}[t]
  \centering
  \caption{
    Ablation study on the components of ETGPO.
    The first ablation replaces taxonomy construction with raw failure sampling.
    The second ablation replaces LLM-generated guidance with direct insertion of error category descriptions.
    The third ablation compares short guidance with detailed guidance containing fuller explanations
    and examples.
    Overall, each ablation results in a reduction in average performance, indicating that all
    components contribute to the effectiveness of ETGPO.
  }
  \begin{tabular}{@{} lcccc @{}}
\toprule
\textbf{Method} & \textbf{AIME} & \textbf{FOLIO} & \textbf{Avg.} & $\boldsymbol{\Delta}$ \\
\midrule

ETGPO
& 53.72 ± 0.28
& 79.53 ± 0.53
& 66.63
& -- \\

Ablation 1: Raw Sampling
& 53.39 ± 0.84
& 77.13 ± 0.45
& 65.26
& -1.37 \\

Ablation 2: Direct Categories
& 52.52 ± 1.27
& 78.33 ± 0.96
& 65.43
& -1.20 \\

Ablation 3: Short Guidance
& 52.20 ± 0.96
& 76.62 ± 0.90
& 64.41
& -2.22 \\

\bottomrule
\end{tabular}

  \label{t:ablations}
\end{table*}

\subsection{Detailed Analysis}
We next present a set of analysis experiments designed to study the impact of
various hyperparameters.
All experiments are conducted on the validation sets of AIME and FOLIO, using
GPT-4.1-mini as the backbone model and GPT-4.1 as the optimizer model.
For each configuration, we run each compared method three times, with 20
evaluation runs per generated prompt.
Unless otherwise specified, all other hyperparameters are fixed to their default
values.
We also include the base Chain of Thought prompt as a baseline in these analyses.
Note that an experiment with \(G\), which determines the number of top error
categories selected for guidance generation, can be found in \cref{a:hyperparameter_g}.

\paragraph{Number of error collection runs}
In the first experiment, shown in \cref{t:hyperparameter_k}, we study the effect
of the parameter \(K\), which controls the number of error collection runs, on
the accuracy of the final prompt.
The effect of \(K\) varies across datasets.
For AIME, accuracy improves as \(K\) increases from 1 to 5, but saturates at
\(K = 10\).
For FOLIO, the best performance is achieved already at \(K = 1\), with accuracy
saturating for larger values.
This suggests that additional execution traces provide diminishing returns for
FOLIO, likely because multiple runs of the same problem do not yield
substantially different error patterns.

\paragraph{Error reduction}
In the second experiment, shown in \cref{fig:failure_reduction}, we examine whether
the generated guidance effectively reduces the types of errors captured by the
different error categories.
Using the optimizer LLM for categorization, we compare the number of failed traces
assigned to the top six error categories before optimization (baseline) and after
optimization, evaluated on the validation sets of AIME and FOLIO, respectively.
As shown in the bar charts, the number of failed traces decreases consistently
across all error categories after optimization, demonstrating the effectiveness
of the prompts generated by our method in mitigating these errors.

\subsection{Ablation Experiments}
We also conduct a set of ablation experiments to assess the contribution of
different components of our method.
The results are reported in \cref{t:ablations}.
Similar to the analysis experiments, all ablations are conducted on the
validation sets of AIME and FOLIO, using GPT-4.1-mini as the backbone model and
GPT-4.1 as the optimizer model.
For each ablation setting, we run the method three times, with 20 evaluation runs
per generated prompt.

\paragraph{Error taxonomy creation}
The first ablation studies the importance of the error taxonomy creation step.
Specifically, it examines whether explicitly categorizing errors across failed
traces provides additional value over directly using raw failed traces.
In this ablation, we skip taxonomy creation and instead pass a sample of 10 failed
traces (one per problem) directly to the guidance generation step.
The results indicate that error taxonomy creation is beneficial on average, with
performance on FOLIO dropping by approximately 2 percentage points when this step
is removed.
In contrast, the drop isn't as significant on AIME, suggesting that a small
sample of failed traces might be sufficient to cover the most salient error patterns
for this dataset.

\paragraph{Guidance generation}
The second ablation evaluates the role of the guidance generation step itself.
Rather than generating guidance text, this ablation directly constructs the final
prompt by inserting the selected error category descriptions as bullet points,
preceded by a generic preamble encouraging the model to watch for common error
patterns.
The results show a clear performance degradation across both datasets, indicating
that explicit guidance generation adds value beyond error category descriptions alone.
In particular, the guidance generation step contributes actionable advice and
concrete correct and incorrect examples that help the model better understand
when and how each error category applies.

\paragraph{Detailed vs short guidance}
The third ablation examines the effect of guidance granularity.
We test a variant of the final prompt in which each error category is summarized
using only one to two lines, rather than a more detailed description.
As shown in the results, this ablation yields the largest performance drop across
both datasets, with an average decrease of approximately two percentage points.
This finding highlights the importance of providing descriptive guidance
for each error category in the final prompt.

\section{Conclusion}
\label{s:conclusion}

In this work, we proposed \textbf{Error Taxonomy-Guided Prompt
Optimization (ETGPO)}, a feedback-driven APO method that builds an error
taxonomy to provide an interpretable view of the dataset-level error
landscape and generates guidance only for the most prevalent error categories.
By focusing on highly prevalent errors, we ensured that prompt edits
generalize beyond the validation set. This allowed us to bypass expensive
iterative validation, making our method substantially more efficient,
as demonstrated in \cref{s:evaluation}. Despite this efficiency,
ETGPO remained highly competitive with state-of-the-art methods,
outperforming on multiple datasets across mathematical reasoning,
question answering, and logical reasoning.



\bibliography{etgpo}
\bibliographystyle{icml2026}

\newpage
\appendix
\onecolumn

\section{Experiment on number of top error categories}
\label{a:hyperparameter_g}
In this experiment, shown in \cref{t:hyperparameter_g}, we analyze the
effect of the parameter \(G\), which determines the number of top error
categories selected for guidance generation.
For AIME, accuracy improves steadily as \(G\) increases, reflecting the presence
of more than ten informative error categories.
In contrast, performance on FOLIO saturates after \(G = 5\).
This behavior is likely due to the fact that FOLIO typically yields around ten
error categories, and categories beyond the top five contribute limited
additional signal for learning effectiveness.

\section{Error Taxonomy Creation Prompts}
\label{a:appendix_1}
Error taxonomy creation uses two prompts to categorize errors found in failed
traces. The first prompt, shown in \cref{fig:prompt-taxonomy-first-1}
and \cref{fig:prompt-taxonomy-first-2}, categorizes the first batch of 
traces. The second prompt, shown in \cref{fig:prompt-taxonomy-subsequent-1}
and \cref{fig:prompt-taxonomy-subsequent-2}, also contains the error categories
discovered so far and prompts the LLM to categorize into existing categories,
only creating new categories if substantially different errors are encountered.
Note that both prompts are split across figures due to their length.

\section{Guidance Generation Prompt}
\label{a:appendix_2}
\cref{fig:prompt-guidance-1} and \cref{fig:prompt-guidance-2} contain the
prompt used for guidance generation. Again, note that the prompt
is split across figures due to its length.

\section{ETGPO-Optimized Prompts}
\label{a:appendix_3}
We present prompts optimized by ETGPO for the datasets HMMT, MMLU-Pro, Musique,
AR-LSAT, and FOLIO in \cref{fig:hmmt_prompt}, \cref{fig:mmlupro_prompt},
\cref{fig:musique_prompt}, \cref{fig:ar_lsat_prompt}, and \cref{fig:folio_prompt}
respectively.

\begin{table}[t]
  \centering
  \caption{
    Effect of varying \(G\), the number of top error categories selected for
    guidance generation, on final prompt accuracy for the AIME and FOLIO
    validation sets.
    Each dataset's accuracy saturates at a different \(G\), suggesting the
    optimal number of error categories varies by task.
  }
  \begin{tabular}{@{} lcc @{}}
\toprule
\textbf{G} & \bf AIME & \bf FOLIO \\
\midrule

Baseline
& 52.35 ± 0.91
& 72.50 ± 0.08 \\

3
& 52.15 ± 0.50
& 78.35 ± 0.60 \\

5
& \underline{52.83 ± 0.12}
& \textbf{79.72 ± 0.21} \\

10
& \textbf{53.72 ± 0.28}
& \underline{79.53 ± 0.53} \\

\bottomrule
\end{tabular}

  \label{t:hyperparameter_g}
\end{table}

\begin{figure}[ht]
\centering
\begin{lstlisting}[style=promptstyle]
You are an expert at analyzing why language models fail on {domain_description}.

## Failure 1
## Failure (Problem {problem_idx}, Run {run_id})

### Problem
{problem}

### Correct Answer
{correct_answer}

### Model's Reasoning
{reasoning}

### Model's Answer
{predicted_answer}

---

## Failure 2
## Failure (Problem {problem_idx}, Run {run_id})
...

## Your Task

Analyze each failure and identify the root cause of each error. Be as descriptive as possible.

For each failure, find:
1. The EARLIEST point in the reasoning where something went wrong
2. What specifically went wrong (calculation error, wrong approach, misunderstanding, etc.)
3. Why this error led to the wrong final answer

Create issue categories that capture each type of error. Categories should be general enough to potentially apply to other traces, but specific enough to be meaningful.

IMPORTANT: Each category must be SELF-CONTAINED and understandable by someone who has NOT seen the original problems.

## Output Format

Return a JSON object with:
```json
{
    "categories": [
        {
            "category_name": "Short descriptive name for this type of error",
            "summary": "One sentence describing the core error pattern.",
            "description": "Detailed description of what goes wrong in this category. Be very specific.",
            "example": "A concrete, self-contained example. Format: 'Problem: [simple problem]. Error: [what the model does wrong]. Correct: [what should happen].'",
            "error_type": "Type of error (e.g., Calculation Error, Wrong Approach, Conceptual Misunderstanding, Missing Step, Logical Fallacy, Factual Error, Incomplete Reasoning, Misreading the Problem)",
            "why_leads_to_wrong_answer": "Explanation of how this error causes wrong answers"
        }
    ],
\end{lstlisting}
\caption{Error taxonomy creation prompt for first batch (Part 1 of 2)}
\label{fig:prompt-taxonomy-first-1}
\end{figure}

\begin{figure}[ht]
\centering
\begin{lstlisting}[style=promptstyle]
    "failure_assignments": [
        {
            "failure_id": 1,
            "problem_idx": <problem_idx>,
            "run_id": <run_id>,
            "category_name": "Name of the category this failure belongs to",
            "trace_details": {
                "trace_specific_location": "Where in the reasoning the error occurred",
                "trace_specific_details": "Specific details about what went wrong"
            }
        }
    ]
}
```
\end{lstlisting}
\caption{Error taxonomy creation prompt for first batch (Part 2 of 2)}
\label{fig:prompt-taxonomy-first-2}
\end{figure}

\begin{figure}[ht]
\centering
\begin{lstlisting}[style=promptstyle]
You are an expert at analyzing why language models fail on {domain_description}.

## Existing Issue Categories

### Category: {category_name}
- Summary: {summary}
- Description: {description}
- Example: {example}
- Error Type: {error_type}
- Why it leads to wrong answer: {why_leads_to_wrong_answer}
- Traces with this issue so far: {trace_count}

### Category: {category_name}
...

## Failure 1
## Failure (Problem {problem_idx}, Run {run_id})

### Problem
{problem}

### Correct Answer
{correct_answer}

### Model's Reasoning
{reasoning}

### Model's Answer
{predicted_answer}

---

## Failure 2
## Failure (Problem {problem_idx}, Run {run_id})
...

## Your Task

For each failure:
1. Determine if the error fits one of the EXISTING categories
2. OR create a NEW category if the error is fundamentally different
\end{lstlisting}
\caption{Error taxonomy creation prompt for subsequent batches (Part 1 of 2)}
\label{fig:prompt-taxonomy-subsequent-1}
\end{figure}

\begin{figure}[ht]
\centering
\begin{lstlisting}[style=promptstyle]
## Output Format

Return a JSON object with:
```json
{
    "new_categories": [
        {
            "category_name": "Short descriptive name for NEW error type",
            "summary": "One sentence describing the core error pattern.",
            "description": "Detailed description of what goes wrong.",
            "example": "A concrete example.",
            "error_type": "Type of error (e.g., Calculation Error, Wrong Approach, Conceptual Misunderstanding, Missing Step, Logical Fallacy, Factual Error, Incomplete Reasoning, Misreading the Problem)",
            "why_leads_to_wrong_answer": "Explanation of how this error causes wrong answers"
        }
    ],
    "failure_assignments": [
        {
            "failure_id": 1,
            "problem_idx": <problem_idx>,
            "run_id": <run_id>,
            "is_new_category": false,
            "category_name": "Name of existing or new category",
            "trace_details": {
                "trace_specific_location": "Where in the reasoning the error occurred",
                "trace_specific_details": "Specific details about what went wrong"
            }
        }
    ]
}
```

Note: "new_categories" should only contain categories that don't exist yet.
\end{lstlisting}
\caption{Error taxonomy creation prompt for subsequent batches (Part 2 of 2)}
\label{fig:prompt-taxonomy-subsequent-2}
\end{figure}

\begin{figure}[ht]
\centering
\begin{lstlisting}[style=promptstyle]
You are an expert at improving language model performance on {domain_description}.

I have identified the following error categories from model failures. Generate guidance to help avoid these errors.

## Category 1: {category_name}

**Statistics:** {failure_count} failures ({coverage_pct}%), {problem_count} problems

**Summary:** {summary}

**Description:** {description}

**Example:** {example}

**Error Type:** {error_type}

**Why it leads to wrong answer:** {why_leads_to_wrong_answer}

---
\end{lstlisting}
\caption{Guidance generation prompt (Part 1 of 2)}
\label{fig:prompt-guidance-1}
\end{figure}

\begin{figure}[ht]
\centering
\begin{lstlisting}[style=promptstyle]
## Category 2: {category_name}
...

## Your Task

Generate guidance text that:
1. Addresses each failure category with specific, actionable advice
2. Is written as instructions TO the model
3. Uses concrete examples where helpful
4. Is prioritized by frequency

Generate DETAILED guidance with examples. Each item should include:
- Description of the error pattern
- Actionable advice on how to avoid it
- WRONG example showing the error
- CORRECT example showing proper approach

## Critical Constraints

- The goal is ACCURACY, not caution. Never generate guidance that encourages the model to refuse, abstain, or say "not specified" when an answer can be reasonably provided.
- CORRECT examples must always show the model providing a substantive answer. Never show abstention/refusal as the correct behavior.

## Output Format

Return a JSON object with:
```json
{
    "guidance_items": [
        {
            "category_name": "Name of the category",
            "guidance_text": "The full guidance text for this category"
        }
    ],
    "preamble": "1-2 sentence introduction",
    "full_prompt": "Complete enhanced prompt starting with base instruction"
}
```

The "full_prompt" should start with:
"Please think step by step and then solve the task."

Then add your preamble and guidance items.
\end{lstlisting}
\caption{Guidance generation prompt (Part 2 of 2)}
\label{fig:prompt-guidance-2}
\end{figure}

\begin{figure}[ht]
\centering
\input{figures/hmmt_prompt}
\caption{
    ETGPO-optimized prompt for the HMMT dataset
    using GPT-4.1-mini (backbone) and GPT-4.1 (optimizer). Truncated.
}
\label{fig:hmmt_prompt}
\end{figure}

\begin{figure}[ht]
\centering
\input{figures/mmlupro_prompt}
\caption{
    ETGPO-optimized prompt for the MMLU-Pro dataset
    using GPT-4.1-mini (backbone) and GPT-4.1 (optimizer). Truncated.
}
\label{fig:mmlupro_prompt}
\end{figure}

\begin{figure}[ht]
\centering
\input{figures/musique_prompt}
\caption{
    ETGPO-optimized prompt for the Musique dataset
    using GPT-4.1-mini (backbone) and GPT-4.1 (optimizer). Truncated.
}
\label{fig:musique_prompt}
\end{figure}

\begin{figure}[ht]
\centering
\input{figures/ar_lsat_prompt}
\caption{
    ETGPO-optimized prompt for the AR-LSAT dataset
    using GPT-4.1-mini (backbone) and GPT-4.1 (optimizer).
}
\label{fig:ar_lsat_prompt}
\end{figure}

\begin{figure}[ht]
\centering
\input{figures/folio_prompt}
\caption{
    ETGPO-optimized prompt for the FOLIO dataset
    using GPT-4.1-mini (backbone) and GPT-4.1 (optimizer). Truncated.
}
\label{fig:folio_prompt}
\end{figure}

\end{document}